\documentclass[letterpaper, 10 pt, conference]{ieeeconf} 
\usepackage[utf8]{inputenc}
\usepackage[
backend=bibtex8, 
style=ieee, 
sorting=none, 
natbib=true, 
doi=false, 
isbn=false, 
url=false, 
eprint=false, 
maxcitenames=1, 
mincitenames=1
]{biblatex}
\usepackage{graphicx}
\usepackage{amsmath}
\usepackage[nolist,nohyperlinks]{acronym}
\usepackage{xcolor}
\usepackage[normalem]{ulem}

\newcommand{\specialcell}[2][l]{%
  \begin{tabular}[#1]{@{}l@{}}#2\end{tabular}}

\renewcommand{\bibfont}{\footnotesize}
\usepackage{array}
\usepackage{multirow}
\newcommand{\PreserveBackslash}[1]{\let\temp=\\#1\let\\=\temp}
\newcolumntype{C}[1]{>{\PreserveBackslash\centering}p{#1}}
\newcolumntype{R}[1]{>{\PreserveBackslash\raggedleft}p{#1}}
\newcolumntype{L}[1]{>{\PreserveBackslash\raggedright}p{#1}}

\IEEEoverridecommandlockouts                              %
\acrodef{GQ-CNN}[GQ-CNN]{Grasp Quality CNN}
\acrodef{CNN}[CNN]{Convolutional Neural Network}
\acrodef{GQ-STN}[GQ-STN]{Grasp Quality Spatial Transformer Network}
\acrodef{STN}[STN]{Spatial Transformer Network}
\acrodef{CGD}[CGD]{Cornell Grasping Dataset}
\acrodef{RPN}[RPN]{Region Proposal Network}
\acrodef{FCNN}[FCNN]{Fully Convolutional Network}
\acrodef{IoU}[IoU]{Intersection-over-Union}

\DeclareMathOperator{\atantwo}{atan2}

\title{\LARGE \bf
  GQ-STN: Optimizing One-Shot Grasp Detection\\based on Robustness Classifier
}

\author{Alexandre Gariépy, Jean-Christophe Ruel, Brahim Chaib-draa and Philippe Giguère}

\begin{document}

\maketitle
\thispagestyle{empty}
\pagestyle{empty}

\begin{abstract}
  Grasping is a fundamental robotic task needed for the deployment of household
  robots or furthering warehouse automation. However, few approaches are able to
  perform grasp detection in real time (frame rate). To this effect, we present
  \acf{GQ-STN}, a one-shot grasp detection network. Being based on the
  \acf{STN}, it produces not only a grasp configuration, but also directly
  outputs a depth image centered at this configuration. By connecting our
  architecture to an externally-trained grasp robustness evaluation network, we
  can train efficiently to satisfy a robustness metric via the backpropagation
  of the gradient emanating from the evaluation network. This removes the
  difficulty of training detection networks on sparsely annotated databases, a
  common issue in grasping. We further propose to use this robustness classifier
  to compare approaches, being more reliable than the traditional rectangle
  metric. Our \ac{GQ-STN} is able to detect robust grasps on the depth images of
  the Dex-Net 2.0 dataset with 92.4 $\%$ accuracy in a single pass of the
  network. We finally demonstrate in a physical benchmark that our method can
  propose robust grasps more often than previous sampling-based methods, while
  being more than 60 times faster.
\end{abstract}

\section{INTRODUCTION}
Grasping, corresponding to the task of grabbing an object initially resting on a
surface with a robotic gripper, is one of the most fundamental problems in
robotics. Its importance is due to the pervasiveness of operations required to
seize objects in an environment, in order to accomplish a meaningful task. For
instance, manufacturing systems often perform pick-and-place, but rely on
techniques such as template matching to locate pre-defined grasping
points~\cite{mercier18_learn_objec_local_pose_estim}. In a more open context
such as household assistance, where objects vary in shape and appearance, we are
still far from a completely satisfying solution. Indeed, in an automated
warehouse, it is often one of the few tasks still performed by
humans~\cite{Correll2016}.

To perform autonomous grasping, the first step is to take a sensory input, such
as an image, and produce a grasp configuration. The arrival of active 3D
cameras, such as the \emph{Microsoft Kinect}, enriched the sensing capabilities
of robotic systems. One could then use analytical methods~\cite{Bohg2013} to
identify grasp locations, but these often assume that we already have a model.
They also tend to perform poorly in the face of sensing noise. Instead, recent
methods have explored data-driven approaches. Although sparse coding has been
used \cite{Trottier}, the vast majority of new data-driven grasping approaches
employ machine learning, more specifically deep learning
\cite{zhou18_fully_convol_grasp_detec_networ,park18_class_based_grasp_detec_using,chu2018,Chen2017,
  trottier_resnet}. %
A major drawback to this is that deep learning approaches require
a significant amount of training data. Currently, grasping training databases
based on real data are scant, and generally tailored to specific robotic
hardware~\cite{Pinto2016,Levine2016}. Given this issue, others have explored the
use of simulated data~\cite{Mahler2017,
  bousmalis17:_using_simul_domain_adapt_improv}.

Similarly to computer vision, data-driven approaches in grasping can be
categorized into \emph{classification} and \emph{detection} methods. In
classification, a network is trained to predict if the sensory input (a cropped
and rotated part of the image) corresponds to a successful grasp location. For
the detection case, the network outputs directly the best grasp configuration
for the whole input image. One issue with classification-based approaches is
that they require a search on the input image, in order to find the best
grasping location. This search can be exhaustive, and thus suffers from the
curse of dimensionality~\cite{Lenz2015}. To speed-up the search, one might use
informed proposals~\cite{Mahler2017,park18_class_based_grasp_detec_using}, in
order to focus on the most promising parts of the input image. This tends to
make the approach relatively slow, depending on the number of proposals to
evaluate.

\begin{figure}[t]
  \centering
  \includegraphics[width=0.7\columnwidth]{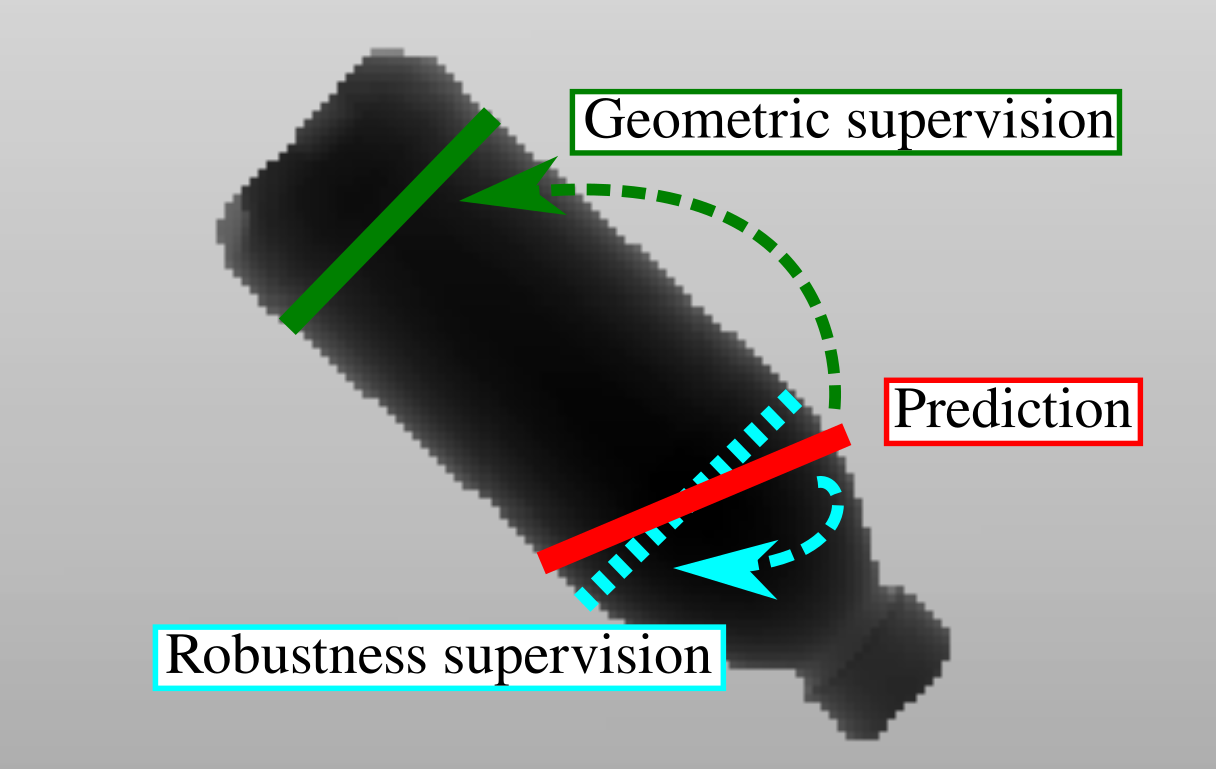}
  \vspace*{-0.3cm}
  \caption{Overview of our method. Classical one-shot methods for grasping
    supervise a prediction (in red) using geometric supervision from
    randomly selected ground truth (in green). We instead suggest to use robustness
    supervision (in cyan) to learn fine-grained adjustments, without requiring a
    ground truth annotation at this specific grasp location.}
  \label{fig:overview}
  \vspace*{-0.6cm}
\end{figure}

While heavily inspired by computer vision techniques, training a network for
detection in a grasping context is significantly trickier. As opposed to classic
vision problems, for which detection targets are well-defined instances of
objects in a scene, grasping configurations are continuous. This means that
there exist a potentially infinite number of successful grasping configurations.
Thus, one cannot exhaustively generate all possible valid grasps in an input
image. Another issue is that grasping databases are not providing the
\emph{absolute best} grasping configuration for a given image of an object, but
rather a (limited) number of \emph{valid} grasping configurations.
In this paper, we propose a one-shot grasping detection architecture for
parallel grippers, based on deep learning. Importantly, our detection approach
on depth images can be trained from \emph{sparse} grasping annotations meant to
train a classifier. As such, it does not require the best grasping location to
be part of the training dataset. To achieve this, we leverage a pre-existing
grasp robustness classifier, called \acf{GQ-CNN}~\cite{Mahler2017}. This is made
possible by the fact that our network architecture directly outputs an image
corresponding to a grasp proposal, allowing it to be fed directly to an
image-based grasp robustness classifier. Our architecture makes extensive use of
the \ac{STN}~\cite{jaderberg15_spatial_trans_networ}, which can learn to perform
geometric transformations in an end-to-end manner. Because our network is based
on \ac{STN}s, the gradient generated by the \ac{GQ-CNN} robustness classifier
will propagate throughout our architecture. Our network is thus able to climb
the robustness gradient, as opposed to simply regressing towards grasp
configurations, which are limited in the training database. In some sense, our
network is able to learn from the implicit knowledge of the quality of a grasp,
knowledge that was captured by \ac{GQ-CNN}.

In short, our contributions are the following:
\begin{enumerate}
\item Describing one of the first techniques to train a one-shot detection
  network on the detection version of the Dex-Net 2.0 dataset. Our network is
  based on an attention mechanism, the \ac{STN}, to perform one-shot grasping
  detection, resulting in our \acf{GQ-STN} architecture;
\item Using the \acf{GQ-CNN} as a supervisor to train this one-shot detection
  network, thus enabling to learn from a limited number of grasp annotations and
  to achieve a high robustness classification score;
\item Showing that our method generalizes well to real-world conditions in a
  physical benchmark, where our \ac{GQ-STN} proposes a high rate of robust grasp.
\end{enumerate}
\section{RELATED WORK}
Over the years, many network architectures have been proposed to solve the
grasping problem. Here, we present them grouped by themes, either
based on their overall method of operation or on the type of generated output.

\subsection{Proposal + Classification Approaches}
Drawing inspiration from previous data-driven methods \cite{Bohg2013}, some
approaches work in a two-stage manner, first by proposing grasp candidates then
by choosing the best one via a classification score. Note that this section does not
include architecture employing \acf{RPN}, as these are applied on a fixed-grid
pattern, and can be trained end-to-end. They are discussed later.

Early work in applying deep learning on the grasping problem employed such a
classification approach. For instance, \textcite{Lenz2015} employed a cascaded
approach of two fully-connected neural networks. The first one was designed to
be small and fast to evaluate and perform the exhaustive search. The second and
larger network then evaluated the best 100 proposals of the previous network.
This architecture achieved 93.7\% accuracy on the \ac{CGD}.

\textcite{Pinto2016} reduced the search space of grasp proposals by only sampling
grasp locations $x,y$ and cropping a patch of the image around this location. To
find the grasp angle, the author proposed to have 18 outputs, separating the angle
prediction into 18 discrete angles by $10^\circ$ increments.

The \emph{EnsembleNet}~\textcite{asifensemblenet} worked in a different
manner. It trained four distinct networks to propose different grasp
representations (regression grasp, joint regression-classification grasp,
segmentation grasp, and heuristic grasp). Each of these proposals was then
ranked by the \emph{SelectNet}, a grasp robustness predictor trained on grasp
rectangles. %

To alleviate the issue of small training datasets labelled manually,
\textcite{Mahler2017} relied entirely on a simulator setup to generate a large
database of grasp examples called Dex-Net 2.0 (see section
\ref{sec:dex-net_dataset}). Each grasp example was rated using a rule-based
grasp robustness metric named \emph{Robust Ferrari Canny}. By thresholding this
metric, they trained a deep neural network, dubbed Grasp-Quality CNN (GQ-CNN),
to predict grasp success or failure. The GQ-CNN takes as input a $32 \times 32$
depth image centered on the grasp point, taken from a top view to reduce the
dimensionality of the grasp prediction. For grasp detection in an image, they
used an antipodal sampling strategy. This way, 1000 antipodal points on the
object surface were proposed and ranked with \ac{GQ-CNN}. Even though their
system is mostly trained using synthetic data, it performed well in a real-world
setting. For example, it achieves a 93\% success rate on objects seen during the
training time and 80\% success rate on novel objects on a physical
benchmark. %
\textcite{park18_class_based_grasp_detec_using} decomposed the search for grasps
in different steps, using
\ac{STN}s. %
The first \ac{STN} acted as a proposal mechanism, by selecting 4 crops as
candidate grasp locations the image. Then, each of these 4 crops were fed into
a single network, comprising a cascade of two \ac{STN}s: one estimated the grasp
angle and the last STN chose the image’s scaling factor and crop. The latter
crop can be seen as a fine adjustment of the grasping location. The four final
images were then independently fed to a classifier, to find the best one. Each
component, being the STNs and the classifier, were trained on \ac{CGD}
separately using ground truth data and then fine-tuned together. This is a major
distinction from other \emph{Proposal + Classification} approaches, as the
others cannot jointly train the \emph{proposal} and
\emph{classification} sub-systems.

\subsection{Single-shot Approaches}
\subsubsection{Regression Approaches}
To eliminate the need to perform the exhaustive search of grasp configurations,
\textcite{Redmon2015Grasp} proposed the first one-shot detection approach. To
this effect, the authors proposed different CNN architectures, in which they
always used AlexNet\cite{Krizhevsky2012} pretrained on ImageNet as the feature
extractor. To exploit depth, they fed the depth channel from the RGB-D images
into the blue color channel, and fine-tuned. The first architecture, named
\emph{Direct Regression}, directly regressed from the input image the best grasp
rectangle represented by the tuple $\{x, y, width, height, \theta\}$. The second
architecture, \emph{Regression + Classification} added object class prediction
to test its regularization effect. \textcite{Kumra2016} further developed this
one-shot detection approach by employing the more powerful ResNet-50
architecture~\cite{He2015}. They also explored a different strategy to integrate
the depth modality, while seeking to preserve the benefits of ImageNet
pre-training. As a solution, they introduced the \emph{multi-modal grasp}
architecture which separated RGB processing and depth processing in two different
ResNet-50 networks, both pre-trained on ImageNet. Their architecture then
performed late fusion, before the fully connected layers performed direct grasp
regression.

\begin{figure}[t]
  \centering
  \vspace{0.2cm}
  \includegraphics[width=0.32\columnwidth]{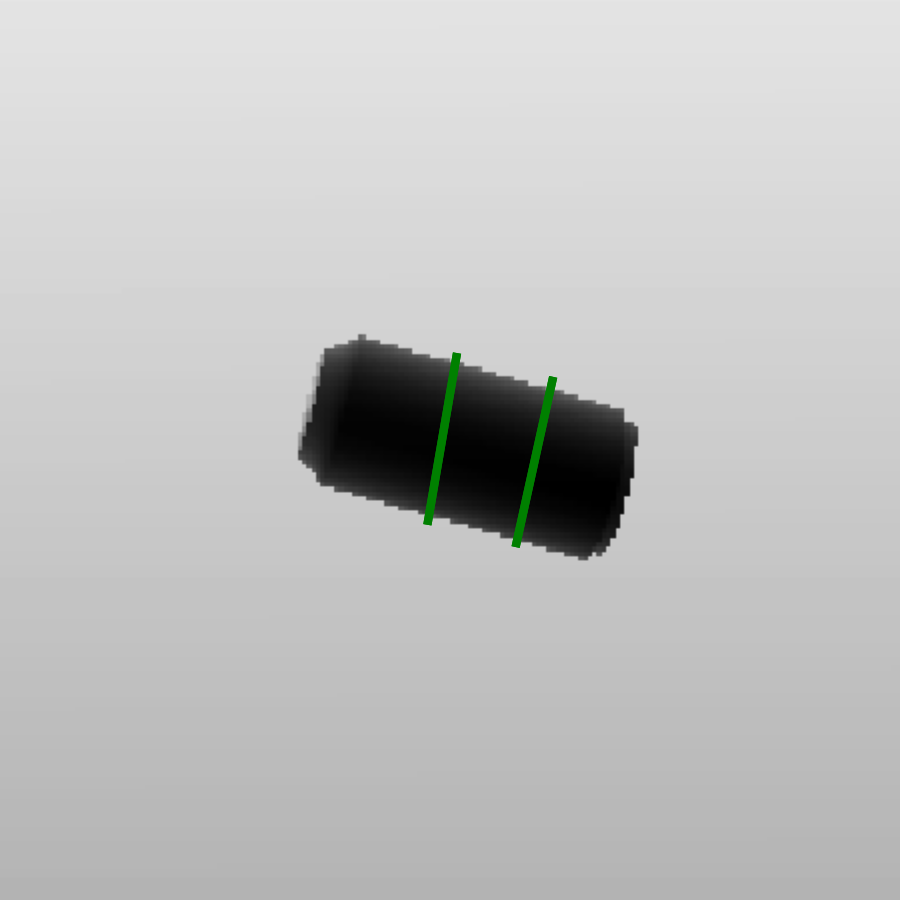}
  \includegraphics[width=0.32\columnwidth]{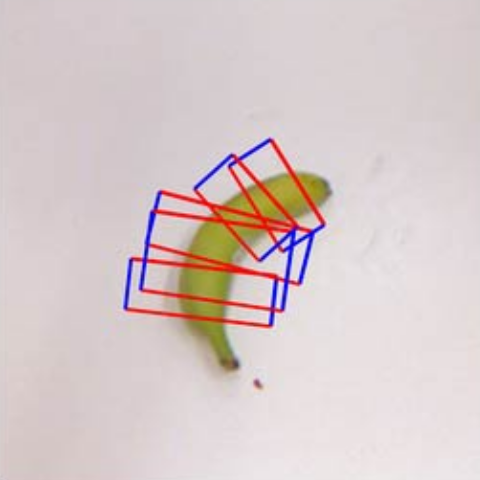}
  \includegraphics[width=0.32\columnwidth]{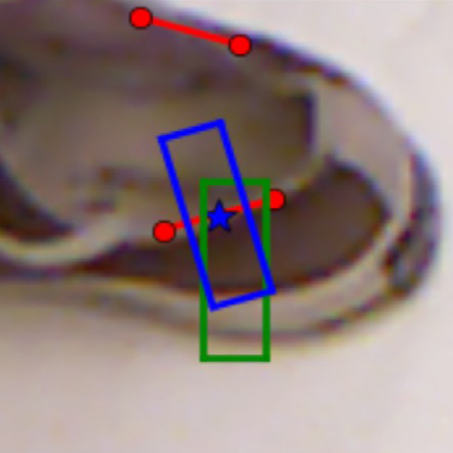}
  \vspace*{-0.2cm}
  \caption{(\textbf{Left}) Training example from the Dex-Net 2.0 detection
    dataset. Notice how there are very few annotations, thus not covering all of
    the possible grasp positions on the entire object. (\textbf{Middle})
    Training example from the \acf{CGD}. These manually-labeled grasp
    annotations tend cover a more important fraction of the object, but for a
    much more limited number of examples. Figure from \cite{Redmon2015Grasp}.\newline
    (\textbf{Right}) Grasp path proposed by \textcite{chen2019convolutional} to
    augment the grasp rectangle representation on the \ac{CGD}. A grasp
    prediction (green) is projected to a grasp path that lies between two
    ground-truth annotation. This allows for better evaluation of detection
    approaches. Figure from \cite{chen2019convolutional}.}
  \label{fig:dexnet_2_dataset}
  \vspace*{-0.6cm}
\end{figure}

\subsubsection{Multibox Approaches}
\textcite{Redmon2015Grasp} also proposed a third architecture,
\emph{MultiGrasp}, separating the image into a regular grid (dubbed Multi-box).
At each grid cell, the network predicted the best grasping rectangle, as well as
the probability of this grasp being positive. The grasp rectangle with the
highest probability was then chosen. \textcite{trottier_resnet} improved results
by employing a custom ResNet architecture for feature extraction. Another
advantage was the reduced need for pre-training on ImageNet.
\textcite{chen2019convolutional} remarked that grasp annotations in grasping
datasets are not exhaustive. Consequently, they developed a method to transform
a series of discrete grasp rectangles to a continuous \emph{grasp path}. Instead
of matching a prediction to the closest ground truth to compute the loss
function, they mapped the prediction to the closest grasp path. This means that
a prediction that falls directly between two annotated ground truths can still
have a low loss value, thus (partially) circumventing the limitations of the
\ac{IoU} metric when used with sparse annotation, as long as the training
dataset is sufficiently densely labeled (see Figure~\ref{fig:dexnet_2_dataset}).
The authors re-used the \emph{MultiGrasp} architecture from
\citeauthor{Redmon2015Grasp} for their experimentation.

\subsubsection{Anchor-box Approaches}
\textcite{zhou18_fully_convol_grasp_detec_networ} introduced the notion of
oriented anchor-box, inspired by YOLO9000~\cite{Redmon2016}. This approach is
similar to MultiGrasp (as the family of YOLO object detectors is a direct
descendant of MultiGrasp~\cite{Redmon2015Grasp}) with the key difference of
predicting offsets to predefined anchor boxes for each grid cell, instead of
directly predicting the best grasp at each cell. \textcite{chu2018} extends
MultiGrasp to multiple object grasp detection by using region-of-interest
pooling layers~\cite{Ren2015}. 

\subsubsection{Discrete Approaches}
\textcite{johns16_deep_learn_grasp_funct_grasp} proposed to use a discretization
of the space with a granularity of 1~\emph{cm} and $30^\circ$. In a single pass
of the network, the model predicts a score at each grid location. Their method
can explicitly account for gripper pose uncertainty. If a grasp configuration
has a high score, but the neighboring configurations on the grid have a low
score, it is probable that a gripper that has a Gaussian error on its position
will fail to grasp at this location. The authors explicitly handled this problem
by smoothing the 3D grid (two spatial axis, one rotation axis) by a Gaussian
kernel corresponding to the gripper error.

\textcite{Satish2019} introduced a fully-convolutional successor to \ac{GQ-CNN}.
It extends \ac{GQ-CNN} to a $k$-class classification where each output is the
probability of a good grasp at the angle $180^{\circ}/k$, similar to
\cite{Pinto2016}. They train their network for this classification task. They
then transform the fully-connected layer into a convolutional layer, enabling
classification at each location of the feature map. This effectively evaluates
each discrete location $x,y$ for graspability.

\section{PROBLEM DESCRIPTION}
\subsection{One-shot Grasp Detection}
Given the depth image of an object on a flat surface, we want to find a grasp
configuration that maximizes the probability of lifting the object with a
parallel-plate gripper. We aimed at performing this detection in a one-shot
manner, i.e. with a single pass of the depth image through our network. As
prediction output, we used the 5D grasp representation $\{x, y, z, \theta, w$\},
where $x, y, z$ captures the 3D coordinates of the grasp, $\theta$ the angle of
the gripper and $w$ its opening. This representation considers grasps taken
from above the object, perpendicular to the table's surface, as in
\cite{Mahler2017, Redmon2015Grasp}. As our network is trained using both the
dataset and the grasp robustness classifier \ac{GQ-CNN} of Dex-Net
2.0~\cite{Mahler2017}, we detail them below.

\begin{figure*}[h]
  \centering
  \vspace{0.2cm}
  \includegraphics[width=0.80\textwidth]{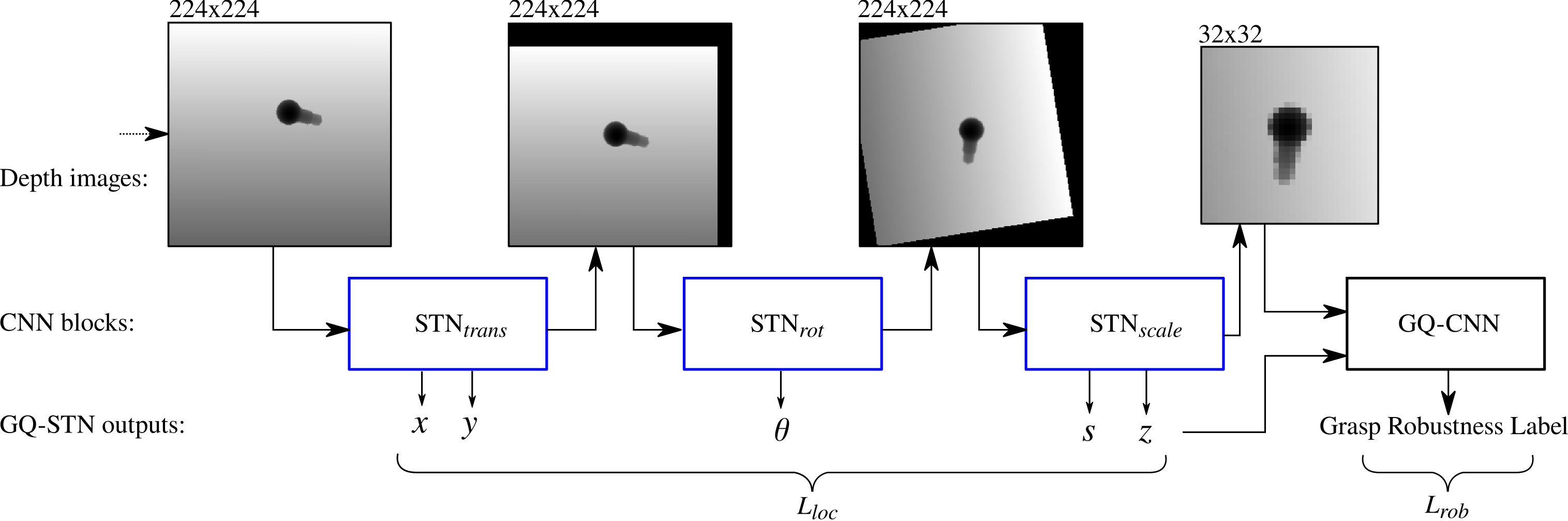}
  \vspace{-0.3cm}
  \caption{Our complete one-shot \ac{STN}-based architecture. The three
    \ac{STN}s learn respectively translation to the grasp's center, rotation to
    the grasp's angle and scaling to the grasp's opening. The intermediary
    outputs of the \ac{STN}s are fully observable and are used to determine the
    grasp location. The last \ac{STN} feeds into \ac{GQ-CNN}, which predicts a
    grasp robustness label. A detailed view of a \ac{STN} block is depicted in
    Fig. \ref{fig:stn_block}.}
  \label{fig:architecture}
  \vspace{-0.6cm}
\end{figure*}

\subsection{Dex-Net 2.0 Dataset}
\label{sec:dex-net_dataset}
Dex-Net 2.0 is a large-scale simulated dataset for parallel-gripper grasping. It
contains 6.7~million grasps on pre-rendered depth images of 3D models. These 3D
models come from two different sources. 1,371 models come from
3DNet~\cite{wohlkinger20123dnet}, a synthetic model dataset built for
classification and pose estimation. The other 129 additional models are laser
scans from KIT~\cite{kasper2012kit}. All of the 3D models were resized to fit
within a 5~\emph{cm} parallel gripper.

The grasp labels in the Dex-net 2.0 dataset were acquired via random sampling of
antipodal grasp candidates. A heuristic-based approach developed in previous
work (Dex-Net 1.0\cite{Mahler2016}) was used to compute a robustness metric.
This metric was thresholded to determine the grasp robustness label, i.e.
\emph{robust} vs. \emph{non-robust}.

Learning one-shot grasp detection on the Dex-Net 2.0 dataset is in itself a
challenging task, because of the few positive annotations per image. Annotations
are very sparse compared to \acf{CGD}, a standard dataset used in one-shot grasp
detection. For instance, it can be seen from Figure \ref{fig:dexnet_2_dataset}
that the ground truth annotation of Dex-Net 2.0 is clearly sparser than \ac{CGD}.
This prevents the grasp annotation augmentations method such as grasp
path~\cite{chen2019convolutional} from being employed on the former.

There are two available versions of the Dex-net 2.0 dataset. The first
version is a classification dataset. It was used by \textcite{Mahler2017} to train
\ac{GQ-CNN} . It contains $32 \times 32$ depth images of grasp candidates with
associated grasp robustness metrics, which are thresholded to obtain robustness
labels. The authors also released a detection version of the dataset. This
version contains the centered depth images of the object, at full resolution
($400 \times 400$).

Please note that in this work, we used the original Dex-Net 2.0 annotations.
Recently published work \cite{Satish2019} developed a sampling method for
generating additional annotations for the Dex-Net 2.0 images. Our approach could
potentially benefit from more detection annotations on images contained in the
Dex-Net 2.0 dataset. Still, for a given object, there is an infinity of possible
grasp configurations which cannot all be annotated. Instead of improving
learning at the annotation level, our approach, described in the following section,
explicitly handles this inherent constraint.

\section{\ac{GQ-STN} NETWORK ARCHITECTURE}

In this paper, we propose \acf{GQ-STN}, a neural network architecture for one-shot grasp
detection based on the \acf{STN}. This architecture enables us to train directly
on a robustness label outputted by \ac{GQ-CNN}, unlike previous one-shot grasp
detection methods that enforce robustness implicitly through geometric
regression on annotated locations. 

\subsection{Spatial Transformer Network}
\begin{figure}[t]
  \centering
  \vspace{0.2cm}
  \includegraphics[width=\columnwidth]{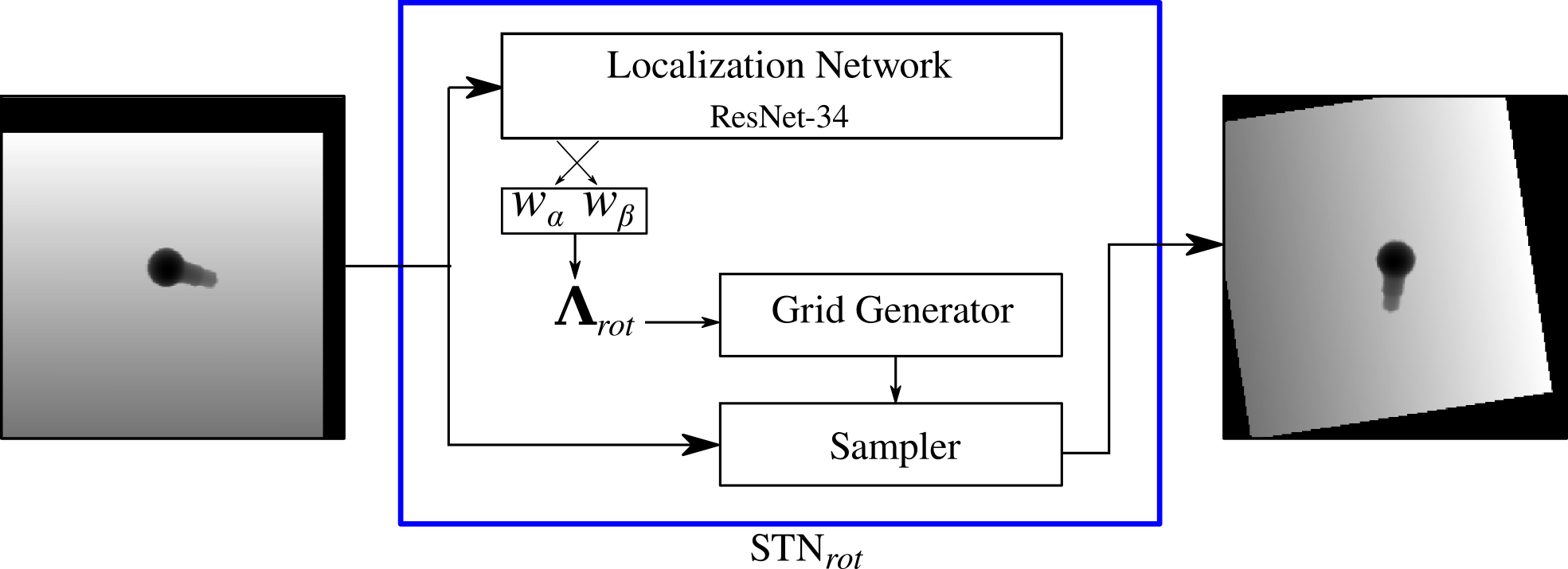}
  \vspace*{-0.7cm}
  \caption{A \acf{STN} block performing a rotation of $\theta$ on an input depth
    image, aligning the image to the grasp's axis. A ResNet-34 localization
    network predicts the transformation matrix $\Lambda_{rot}$. This is the
    second of the three \ac{STN}s shown in Fig. \ref{fig:architecture}.}
  \label{fig:stn_block}
  \vspace*{-0.6cm}
\end{figure}

The main component in our single-shot detection architecture is the
\acf{STN}~\cite{jaderberg15_spatial_trans_networ}. In some sense, it acts as an
attention mechanism, by narrowing/reorienting objects in a more canonical
representation for the task at hand. It is a drop-in block that can be inserted
between two feature maps of a \ac{CNN} to learn a spatial transformation of the
input feature map. The \acf{STN} consists of three parts: a \emph{localization
  network}, a \emph{grid generator} and a \emph{sampler}. The localization
network learns a transformation matrix $\Lambda^{2\times 3}$ based on the input
feature map. The grid generator and the sampler transform the input feature map
by the geometric transformation specified by $\Lambda$. It does so in a fully
differentiable manner, in a process similar to texture mapping. It can thus
stretch, rotate, or skew the input feature map, resulting in a new feature map
as output. A pure rotation transformation is illustrated in
Figure~\ref{fig:stn_block}.

A \acf{STN} can be constrained to only represent specific geometric
transformations, instead of freely learning the six elements of $\Lambda$. In
our approach, we will employ three different transformations matrices:
\begin{align*}
  \Lambda_{trans} = &
                   \begin{bmatrix}
                     1 & 0 & x \\
                     0 & 1 & y \\
                   \end{bmatrix},\quad
  \Lambda_{scale} =
                   \begin{bmatrix}
                    s  & 0 & 0 \\
                    0  & s & 0\\
                   \end{bmatrix},\\
  &\Lambda_{rot} =
                   \begin{bmatrix}
                     \cos{\theta} & -\sin{\theta} & 0 \\
                     \sin{\theta} & \cos{\theta} & 0\\
                   \end{bmatrix}.
\end{align*}
$\Lambda_{trans}$ represents a relative translation by a factor of $x, y \in
[-0.5,0.5]$, $\Lambda_{rot}$ a rotation by an angle $\theta$ and
$\Lambda_{scale}$ an isotropic scaling by a factor of $s$.

\subsection{Full architecture}

Instead of predicting all transformations in a single network, we used a cascade
of three \ac{STN} blocks, STN$_{trans}$ STN$_{rot}$ and STN$_{scale}$, which are
respectively constrained by $\Lambda_{trans}$, $\Lambda_{rot}$ and
$\Lambda_{scale}$. In other words, STN$_{trans}$ learns the translation $x, y$
to the grasp center, STN$_{rot}$ learns the rotation $\theta$ of the gripper and
STN$_{scale}$ learns a scaling $s$ representing the opening of the gripper. A
motivation behind this architecture is to isolate the regression of the angle
$\theta$, which is a challenging task for a one-shot network according to
\textcite{park18_rotat_ensem_modul_detec_rotat_invar_featur}. All Spatial
Transformer Networks (STN) were applied directly to the 1-channel depth map;
contrary to \textcite{Kumra2016}, we found no benefit in using a 3-channel
version pre-trained on ImageNet for the \ac{STN}s. All \ac{STN}s also output a
depth image, meaning that the communication between blocks of the network is not
conducted via high-level feature maps, but via fully-observable depth images.

We used ResNet-34 as localization networks in all three \acf{STN}s, as in
\cite{park18_class_based_grasp_detec_using}. This yielded slightly better results
than the smaller ResNet-18 while maintaining a reasonable training time. Drawing
from \cite{Redmon2015Grasp} and \cite{Redmon2016}, the output layers of the
ResNet-34 computed the elements of $\Lambda_{\star}$ as follows:
\begin{align*}
  \begin{split}
  x &= \sigma(w_x) - 0.5\\
  y &= \sigma(w_y) - 0.5\\
  \end{split}
  \hspace{0.5cm}
  \begin{split}
  \alpha &= \sigma(w_{\alpha})\\
  \beta &= \sigma(w_{\beta})\\
  \theta &= \atantwo(\alpha, \beta) / 2\\
  \end{split}
  \hspace{0.5cm}
  \begin{split}
  s &= \gamma e^{w_s}\\
  z &= w_z\\
  \end{split}
\end{align*}
The tuples $\{w_x,w_y\}, \{w_{\alpha}, w_{\beta}\}$ and $\{w_s, w_z\}$ are the raw outputs
of the localization networks of respectively STN$_{trans}$ STN$_{rot}$ and
STN$_{scale}$. To break the two-fold rotational symmetry of the angle
prediction, we predict $\alpha, \beta$ which are respectively the sine and
cosine of twice the angle $\theta$, as in \cite{Redmon2015Grasp}. $\gamma$ is
the mean scaling factor in the training set. In conjunction with the scaling
$s$, the last \ac{STN}'s localization network also predicts the normalized
gripper's height $z$.

The input of the complete network, illustrated in Figure~\ref{fig:architecture},
is a $224 \times 224$ depth image. The translation and rotation \ac{STN}s both
generate a depth image of the same size as the input, while the STN$_{scale}$
generates a depth image at a resolution of $32\times 32$. STN$_{scale}$ is
followed by \ac{GQ-CNN}. The latter predicts a grasp robustness label given the $32
\times 32$ image outputted by STN$_{scale}$. We use pre-trained weights made
available by \citeauthor{Mahler2017} for \ac{GQ-CNN}. These weights are frozen
throughout training. At evaluation time, \ac{GQ-CNN} is not required for grasp
detection. However, because evaluating a single grasp on GQ-CNN is low-cost, we
keep GQ-CNN to avoid a GPU memory transfer cost later if we need a robustness
label associated with a detection.

Note that every block in the architecture is fully differentiable, thus allowing
us to leverage information from the error on the grasp robustness label, by
back-propagating from the latter all the way back to the first \ac{STN}.

\subsection{Training}
At each step of training, we randomly select a ground truth positive grasp
example from the Dex-net 2.0, thus obtaining target values for location
$\Lambda_{trans}^{gt}$, $\Lambda_{rot}^{gt}$ and $\Lambda_{scale}^{gt}$. We
train the network using two types of supervision:
\begin{itemize}
\item \textbf{Localization loss} $L_{loc}$: the $L_2$ loss on the predictions of the localization networks of the \ac{STN}s using $\Lambda_*^{gt}$; 
\item \textbf{Robustness loss} $L_{rob}$: the cross-entropy loss on the output of \ac{GQ-CNN}, where the expected value is a positive grasp label.
\end{itemize}
 The total loss $L_{tot}$ is given by:
\begin{align*}
  L_{tot} &= \xi L_{loc} + (1 - \xi)L_{rob}.
\end{align*}

The training regimen begins with $\xi = 1$ and we gradually slide the loss
mixing parameter toward $\xi = 0$. This way, we bootstrap the learning of our
architecture with groud-truth grasp positions. These provide strong cues to the
\ac{STN}s, via the loss $L_{loc}$. As we reach $\xi = 0$, the network training
then focuses on directly improving the grasp quality metric, irrespective of
grasp positions. Importantly, this allows our one-shot detection network to
learn from sparsely labeled ground-truth, by eventually strictly focusing on a
grasp robustness metric provided by GQ-CNN. The bootstrapping induced by $\xi >
0$ was necessary for the network training to converge, enabling a proper
focus on the object. It can be seen in Figure~\ref{fig:stn_block} that
transformations on the depth image introduce artifacts on the edges. If one
would start training with $\xi=0$, the network would enter a degenerate state
where edge artifacts are mistaken for object edges.

During early stages of bootstrapping when $\xi > 0$, training tend to be quite
unstable. There is an accumulation of error where, for instance, STN$_{scale}$
cannot provide a good prediction because of errors made by STN$_{trans}$ and
STN$_{rot}$, resulting in a high $L_{loc}$. We solved this issue by using a
\emph{teacher forcing} approach \cite{Goodfellow-et-al-2016} where the \ac{STN}s
are trained in a disjoint manner. Instead of using the $\Lambda_{trans}$ and
$\Lambda_{rot}$ predicted by the first and second localization networks
respectively, we directly transform the images using the ground truth
information $\Lambda_{trans}^{gt}$, $\Lambda_{rot}^{gt}$. Teacher forcing allows
the three \ac{STN} to be trained simultaneously, instead of training them in
sequence as proposed in \cite{park18_class_based_grasp_detec_using}, resulting
in a shorter training time. Teacher forcing is disabled after $\xi=0$, allowing
a joint training of all parameters on $L_{rob}$.

\section{EXPERIMENTS AND EVALUATION}
We compared our architecture against three baselines: the single-shot
DirectGrasp and MultiGrasp architectures\cite{Redmon2015Grasp} and the approach
based on \emph{Proposal+Classification} from Dex-Net 2.0\cite{Mahler2017} that
we will refer to as Prop+GQ-CNN. For DirectGrasp and MultiGrasp, we replaced the
AlexNet feature extractor by a ResNet feature extractor, as seen in
\textcite{Kumra2016}. We trained our \ac{GQ-STN} model and both DirectGrasp and
MultiGrasp on $80\%$ of the Dex-Net 2.0 dataset and held $20\%$ in a test set.
For the Prop+GQ-CNN approach, we used the pre-trained model made available by
the authors.

We further tested GQ-STN, DirectGrasp and MultiGrasp on the Jacquard
dataset\cite{depierre18_jacquar}. Note that because Jacquard does not contain
any gripper height information, we could not train the architectures on this
dataset, as the gripper height is a required input of GQ-CNN. Therefore,
Jacquard is only used here for testing networks that were trained on the Dex-Net
2.0 dataset.

We implemented all the architectures using the Tensorflow library. We trained
all models 40 epochs with the Adam Optimizer. For \ac{GQ-STN}, we had the
following scheduling for $\xi$ and the learning rate $lr$: $6$ epochs at
$\xi=1.0,lr=1\times10^{-3}$, $3$ epochs at $\xi=0.5,lr=2\times 10^{-4}$, $3$
epochs at $\xi=0.2,lr=4\times 10^{-5}$, $9$ epochs at $\xi=0.0,lr=4\times
10^{-5}$, and a fine-tuning stage of $19$ epochs at $\xi=0.0,lr=8\times 10^{-6}$
using early stopping. Teacher-forcing was turned on for only the first 12
epochs.

For DirectGrasp and MultiGrasp, we employed the same $lr$ schedule, though they
converged faster that GQ-STN and the last fine-tuning step with
$lr=8\times10^{-6}$ did not improve results. We kept the models that had the
highest rectangle metric score in validation (see
Section~\ref{sec:rectangle_eval}). We had for all models a $L_2$ regularization
factor of $1\times10^{-7}$.

We compared the quality of predictions of the single-shot baselines and our
\ac{GQ-STN} network using the robustness classification metric
(Sec.~\ref{sec:robust_eval}). We also evaluated these three models according to
the rectangle metric (Sec.~\ref{sec:rectangle_eval}). Finally, we conducted real
world grasping experiments (Sec.~\ref{sec:robot_eval}) where we evaluated
MultiGrasp, our GQ-STN, and Prop+GQ-CNN. All experiments and training were
conducted on a Desktop computer with a 4~\emph{GHz} Intel i7-6700k and an NVIDIA
Titan X GPU.

\subsection{Robustness Classification via GQ-CNN}
\label{sec:robust_eval}

\begin{figure}[t]
  \vspace{0.2cm}
  \centering
  \includegraphics[width=0.49\columnwidth]{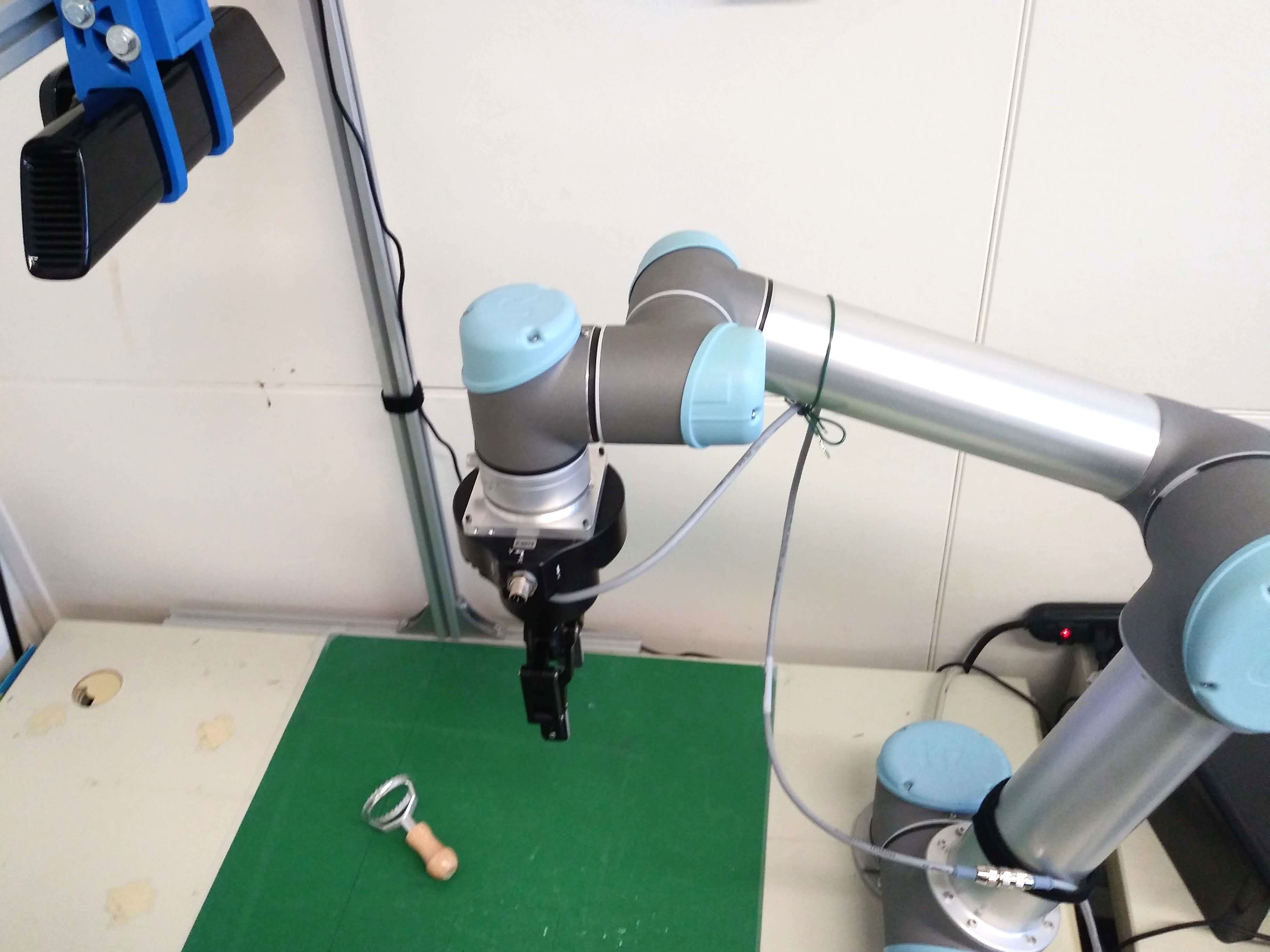}
  \includegraphics[width=0.49\columnwidth]{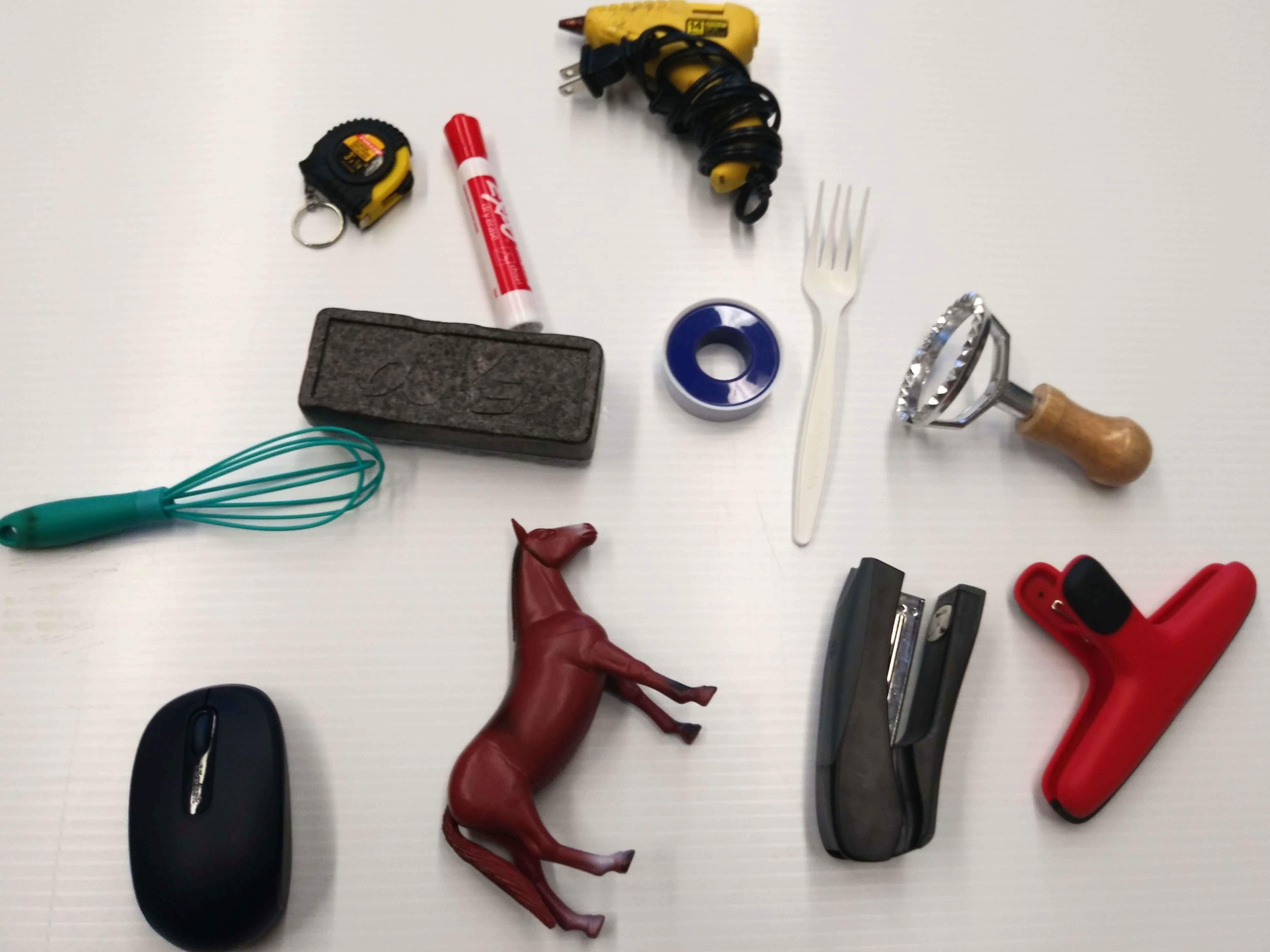}
  \vspace*{-0.6cm}
  \caption{(\textbf{Left}) Physical setup used for evaluation. It contains a UR5 arm, a
    Robotiq 85 gripper and a Microsoft Kinect sensor. (\textbf{Right}) Set of 12
    household and office objects used in tests.}
  \label{fig:physical_benchmark}
  \vspace*{-0.6cm}
\end{figure}

\textcite{asifensemblenet} used \emph{SelectNet}, a CNN trained for grasp
evaluation. However, SelectNet was trained based on a metric similar to Jaccard,
which is problematic (see Sec.~\ref{sec:rectangle_eval}) and would thus provide
for poor evaluation. In our situation, we preferred instead to use the
pre-trained classifier \ac{GQ-CNN}~\cite{Mahler2017} for robustness evaluation
of predicted grasp configurations. Indeed, this classifier was trained with a
heuristic-based robustness evaluation metric named \emph{Robust Ferrari-Canny}.
Moreover, the \ac{GQ-CNN} was found experimentally to be an excellent predictor
of grasp success, with $94\%$ on known objects and a precision of $100\%$ on
unknown objects~\cite{Mahler2017}. As a reminder, the \ac{GQ-CNN} takes as an
input a $32 \times 32$ depth image centered around the grasp location and
classifies whether or not it is a robust grasp location.

We evaluated our architecture and the one-shot baseline architectures
(DirectGrasp and MultiGrasp) using this robustness evaluation methodology. For
the baselines, we extracted a $32 \times 32$ depth image around the
grasp rectangle and fed it to \ac{GQ-CNN} for classification. The output image
crop generated automatically by our \ac{GQ-STN} architecture was used
directly for evaluation. For all architectures, a grasp configuration was
considered positive if it was classified as robust by \ac{GQ-CNN}. Robustness
classification results are found in Table~\ref{tab:robust_metric}.

\subsection{Rectangle Metric}
\label{sec:rectangle_eval}

\begin{figure}[b]
  \vspace*{-0.4cm}
  \centering
  \includegraphics[width=0.24\columnwidth]{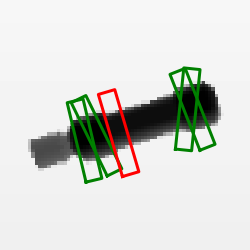}
  \includegraphics[width=0.24\columnwidth]{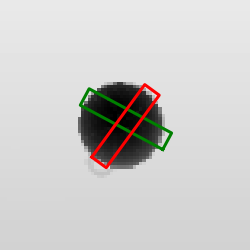}
  \includegraphics[width=0.24\columnwidth]{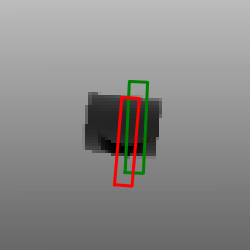}
  \includegraphics[width=0.24\columnwidth]{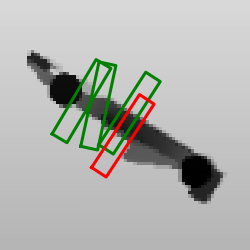}
  $\underbrace{\hspace{0.45\columnwidth}}_{negative}$
  \hspace{0.2cm}
  $\underbrace{\hspace{0.45\columnwidth}}_{positive}$
  \vspace{-0.1cm}
  \caption{Examples of grasp predictions (in red) and ground truth annotations
    (in green) depicting the limitations of the grasp rectangle as an evaluation
    metric. (\textbf{Left}) Examples of negatives grasps of the rectangle metric
    classified robust by \ac{GQ-CNN}. (\textbf{Right}) Examples of positive
    grasps of the rectangle metric classified non-robust by \ac{GQ-CNN}.}
  \label{fig:result_false_pos_neg}
  \vspace{-0.05cm}
\end{figure}

The rectangle metric is a standard evaluation metric for grasping systems
introduced in \cite{Jiang2011}. Given a grasp prediction $P$ and its closest
ground truth $G$, $P$ is considered correct if both:

\begin{enumerate}
\item the angle difference between $P$ and $G$ is below $30^{\circ}$;
\item the Jaccard index $J(P,G) = |P \cap G|/|P \cup G|$ is greater that $0.25$.
\end{enumerate}

Note that the Dex-Net 2.0 dataset does not contain the rectangle height $h$
required by the rectangle grasp representation. We simply assumed that $h = w /
5$, which corresponds to the size of the gripper's finger tips. Our architecture
does not predict $w$ directly, but an analogous scaling factor $s$. We
considered that $w = s / 3$, which corresponds to how grasps are represented in
the Dex-Net 2.0 dataset. All architectures predict a gripper height $z$ in
addition to the 2D grasp configuration. For the rectangle metric evaluation
purposes, this parameter $z$ is ignored.

We evaluate DirectGrasp, MultiGrasp and \ac{GQ-STN} on the rectangle metric.
Table~\ref{tab:robust_metric} shows that DirectGrasp and MultiGrasp perform
slightly better than \ac{GQ-STN} on the rectangle metric. This is understandable
since they were specifically trained for rectangle regression. However, both
networks have a poor Robustness Classification Metric score.

The rectangle metric is known to have a number of issues
\cite{ghazaei18_dealin_with_ambig_robot_grasp, chen2019convolutional}. First and
foremost, the score bears no physical meaning in terms of grasp robustness, as
it is purely computed in the image space. For example, a grasp rectangle can be
considered as valid (high Jaccard index), even if a finger collides with the
object. Second, for a grasp prediction to be evaluated, there needs to be a
ground truth annotation near the exact position of the prediction. In other
words, the validity of a grasp prediction depends on whether or not it was
annotated in the dataset. This is particularly problematic when evaluating grasp
\emph{detection} frameworks, as for a given object, there is an infinity of
possible grasp configurations which cannot all be annotated. In a
\emph{classification} framework, one does not suffer from this issue, since only
labeled examples are used during evaluation.

To observe the lack of correlation between the rectangle metric and grasp
robustness, we conducted an experiment using the Dex-Net 2.0 dataset. We first
examined the quantity of predicted grasp rectangles that are considered positive
by the rectangle metric but are not robust according to the robustness
classification metric of GQ-CNN, described in Sec.~\ref{sec:robust_eval}. These
account for $46.3\%$ and $30.8\%$ of grasps detected by respectively MultiGrasp
and \ac{GQ-STN}. Conversely, we examined the grasps that are considered negative
according to the rectangle metric but are robust according to the robustness
classification metric. These account for $50.7\%$ and $51.9\%$ of grasps
detected by respectively MultiGrasp and \ac{GQ-STN}. These represent grasp
rectangles that would be positive if they were annotated in the dataset.
Examples are shown in Figure \ref{fig:result_false_pos_neg}. These auxiliary
results show that, especially in the context of sparse grasp annotations such as
with the Dex-Net 2.0 dataset, the rectangle metric does not properly represent
the performance of a grasping system. This further motivates our choice of
evaluating with a robustness classification metric.

\subsection{Metric Results}
Table~\ref{tab:robust_metric} shows that overall, on the Dex-Net 2.0 dataset,
our approach is able to return a significantly higher percentage of high-quality
grasps (92.4\%) than the one-shot detection approach based on MultiGrasp
(30.6\%) and DirectGrasp (25.9\%). This large performance gap can be explained
by the fact that our approach enables us to optimize directly on the robustness
classification metric, which is impossible for the two baselines. For all
approaches, the rectangle metric tends to under-estimate the performance, which
is explainable by sparse grasp annotations of the Dex-Net 2.0 dataset, as
discussed in Section~\ref{sec:rectangle_eval}.

We also tested the three models on the Jacquard dataset, which, contrary to
Dex-Net 2.0, contains dense grasp rectangle annotations. As we can see in
Table~\ref{tab:robust_metric}, our GQ-STN returns significantly more robust
grasp (60.4\%) than the best baseline MultiGrasp (34.2\%). This shows a good
generalization of our method, which is also observed in the physical benchmark
(Section~\ref{sec:robot_eval}).

\begin{table}[b]
  \centering
  \caption{Comparison of one-shot methods on evaluation metrics.}
  \label{tab:robust_metric}
  \begin{tabular}{|l|l|c|c|}
    \hline
    \textbf{Test Dataset} & \textbf{Model} & \multicolumn{2}{c|}{\textbf{Precision ($\%$)}} \\
    & & Rectangle & Robust\\
    \hline \hline
    \multirow{3}{*}{Dex-Net 2.0} & DirectGrasp & 48.1 & 25.9 \\\cline{2-4}
                                 & MultiGrasp & \textbf{48.4} & 30.6 \\\cline{2-4}
                                 & GQ-STN (ours) & 46.7 & \textbf{92.4}\\
    \hline
    \multirow{3}{*}{\specialcell{Jacquard \\ (trained on Dex-Net 2.0)}}
                     & DirectGrasp & 67.4 & 32.7 \\\cline{2-4}
                     & MultiGrasp & \textbf{71.8} & 34.2 \\\cline{2-4}
                     & GQ-STN (ours) & 70.8 & \textbf{60.4} \\
    \hline
  \end{tabular}
\end{table}

\subsection{Physical benchmark}
\label{sec:robot_eval}
\begin{figure}[t]
  \centering
  \vspace{0.2cm}
  \includegraphics[width=0.80\columnwidth]{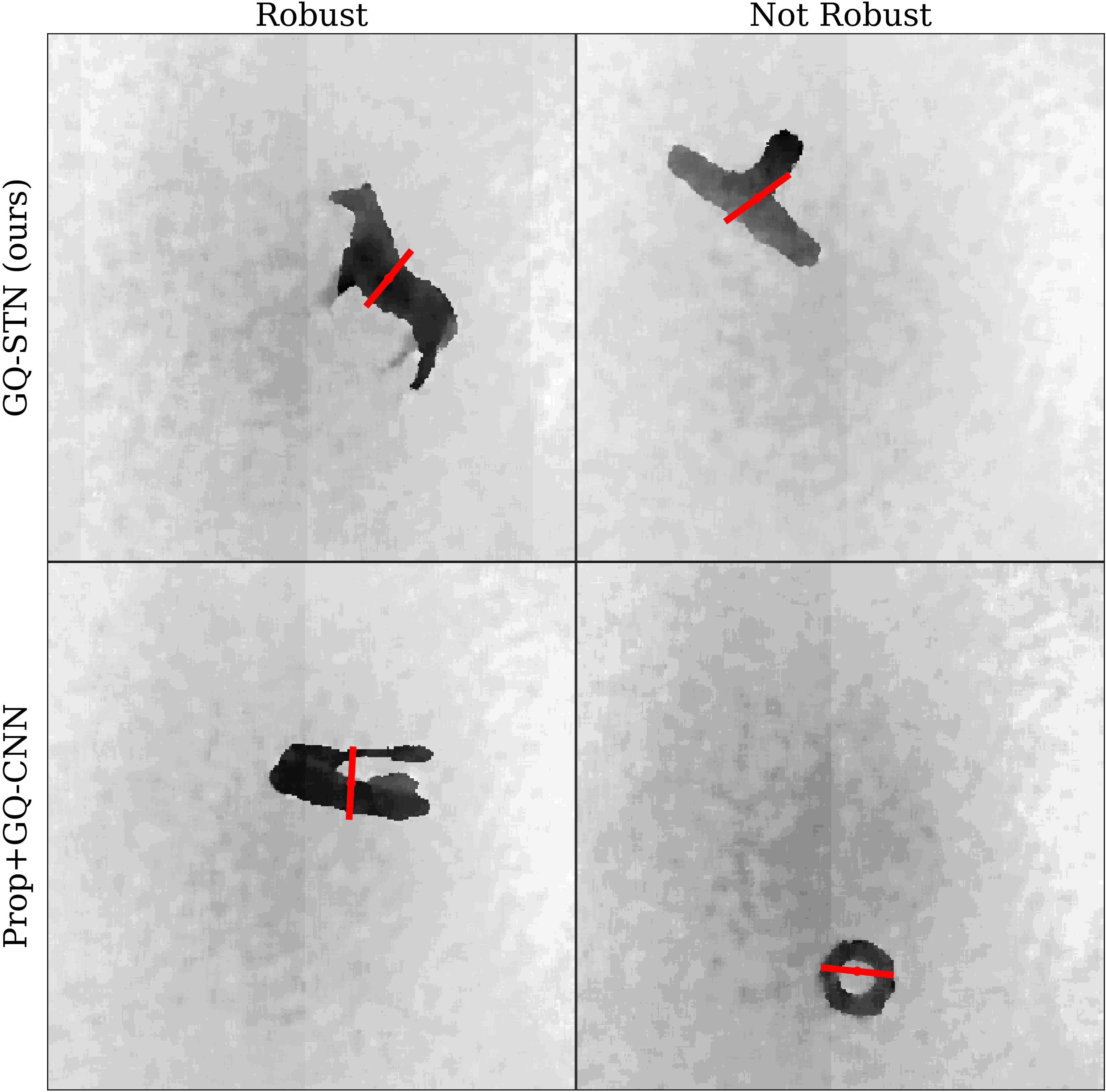}
  \vspace*{-0.2cm}
  \caption{Examples of robust and non-robust grasp detection made by \ac{GQ-STN}
  and Prop+GQ-CNN in our physical benchmark.}
  \label{fig:experimental_results}
  \vspace*{-0.6cm}
\end{figure}

\begin{table}[b]
  \centering
  \caption{Comparison of methods on our physical benchmark.}
  \label{tab:benchmark_results}
  \begin{tabular}{|l|C{1.6cm}|C{1.8cm}|C{1.6cm}|}
    \hline
    \textbf{Model} & \textbf{Success rate ($\%$)} & \textbf{Robust pred. rate ($\%$)} & \textbf{Grasp detect. time (sec)} \\
    \hline \hline
    MultiGrasp & 95 & 21.7 & 0.014\\
    \hline
    GQ-STN (ours) & 96.7 & 61.7 & 0.024\\
    \hline
    \hline
    Prop+GQ-CNN  & 98.3 & 48.3 & 1.5\\
    \hline
  \end{tabular}
\end{table}

We evaluated all three methods in real-world conditions using the physical setup
seen in Figure~\ref{fig:physical_benchmark}. It comprised a \emph{Universal Robots
  UR5} arm, a \emph{Robotiq 85} gripper and a \emph{Microsoft Kinect} sensor.
The \emph{Kinect} sensor was mounted 70~\emph{cm} perpendicular to the table's
surface. Grasp prediction was based on a single rectified depth image, where we
replaced invalid depth pixels using
inpainting~\cite{johns16_deep_learn_grasp_funct_grasp}.

We selected 12 household and office objects for testing, shown in
Figure~\ref{fig:physical_benchmark}. We chose objects that have a good variety
of shape, material and texture and are similar to the one used in
\cite{Mahler2017}. During testing, we placed the target object at a random
position near the center of the table, by shaking it under a box to ensure
random orientation, as in \textcite{Mahler2017}. We then estimated the grasp
configuration with one of the three methods, and used a custom path planner to
execute the grasp motion. The gripper default opening was 8.5~\emph{cm}. It
closed on the object until a maximum force feedback is reached. Upon closure, the
object was lifted from the table and the success evaluated manually. Each of the
12 objects was tested 5 times, for each compared method. In total, we performed
180 grasp attempts.

We computed three metrics in this physical benchmark:
\begin{enumerate}
\item \textbf{Success rate}: Percentage of the lift attempts that resulted in a
  success. We execute the detected grasp even if it is not classified robust by
  the robustness classification metric.
\item \textbf{Robust prediction rate}: Percentage of the time the detected grasp
  (or the top grasp candidate for the sampling-based Prop+GQ-CNN) is robust
  according to the robustness classification metric.
\item \textbf{Grasp detection time}: Time in seconds between capturing an image
  and returning a grasp location. Here, we ignore time taken for inpainting.
\end{enumerate}

As we can see in Tab.~\ref{tab:benchmark_results}, all three methods performed
similarly, within the uncertainty of low samples. However, our method returned a
robust grasp $61.7\%$ of the time, which is significantly more than MultiGrasp
and above Prop+GQ-CNN.

Qualitatively, the approach Prop+GQ-CNN seemed to perform slightly better during
real experiments, especially with larger objects such as the red chips clip. In
some sense, this is not surprising as it evaluated the grasp quality over 1000
positions. Figure~\ref{fig:experimental_results} shows examples of grasp
detection on our physical benchmark. Even though the methods were trained only
on simulated data, its large amount helped generalization to real-world
conditions, as noted as well by \textcite{Mahler2017}. Note that no
domain-randomization was used here, contrary to
\textcite{bousmalis17:_using_simul_domain_adapt_improv}.

In terms of timing, our \ac{GQ-STN} approach is in the same order of magnitude as
the MultiGrasp approach, even though we run an image through three ResNet networks (one per
Localization Network inside the \ac{STN}). The detection time for Prop+GQ-CNN is
two order of magnitudes larger than our approach, i.e. around $60$ times
slower. This limits its ability to perform real-time grasp detection.

Even though \ac{GQ-STN} returns a single grasp and does so much faster,
\ac{GQ-STN} finds a robust grasp more often that Prop+GQ-CNN's sampling.
Considering the high precision of the robustness classification metric, this
enables \ac{GQ-STN} to be used in a framework where we first evaluate the fast
\ac{GQ-STN} then fallback to a slow sampling method if we have not found a
robust grasp, improving the overall average planning time.

\section{CONCLUSION}
In this paper, we present a novel architecture for one-shot
detection of grasp localization, based on the \acf{STN} architecture. With it,
we have demonstrated how one can use supervision from a robustness classifier to
train one-shot grasp detection. On the Dex-Net 2.0 dataset, our method returns
robust grasps more often than a baseline model that is only trained using the
geometric supervision. We showed in a physical benchmark that our method can
find robust grasps in real-world conditions more often that sampling methods,
while still performing real-time (over 40~\emph{Hz}), which is greater than
frame rate grasp detection on a Kinect.

This speed opens up the possibility of carrying out visual servoing for
grasping, for moving objects for instance. If a camera in-hand is used, it makes
it possible to explore a object in real-time, similarly to a next-best-view
approach, akin to \textcite{Levine2016}. There are other interesting research
avenues at the network architecture level for future work. For instance, since
all inputs of the \acf{STN} are similar depth images, one could imagine a
parameter sharing mechanism to speed up the training time and reduce the model
size.

\section*{ACKNOWLEDGEMENTS}
This works was financed by the Fonds de Recherche du Québec – Nature et
technologies and the Natural Sciences and Engineering Research Council
of Canada.

\printbibliography

\end{document}